# Covid-19 Public Sentiment Analysis for Indian Tweets Classification

Mohammad Maksood Akhter, Devpriya Kanojia,

*PDPM Indian Institute of Information Technology, Design and Manufacturing, Jabalpur, India*
Submitted June 2020

## Abstract

When any extraordinary event takes place in the world wide area, it is the social media that acts as the fastest carrier of the news along with the consequences dealt with that event. One can gather much information through social networks regarding the sentiments, behavior, and opinions of the people.
In this paper, we focus mainly on sentiment analysis of twitter data of India which comprises of COVID-19 tweets. We show how Twitter data has been extracted and then run sentimental analysis queries on it. This is helpful to analyze the information in the tweets where opinions are highly unstructured, heterogeneous, and are either positive or negative or neutral in some cases.

## Introduction

With the outbreak of any ubiquitous disease, the social media platform surfs with people's emotions and opinions. It has been shown in studies that these platforms provide a chance for them to be vocal about their thoughts, likeness and also, stating where the problem actually began. With apt research on the behaviors of common people during a crisis is one way to deal with the on-going problem and to assist health professionals by making the relevant data regarding the mental health of the people available to them. Data, which is publicly available, can be extracted to form a dataset. Facebook, Twitter, YouTube, etc can have valued data that consists of the primary issues that are of concern and can be put into use of many professionals. In this research article, we have collected tweets from Twitter, specifically the tweets made in India for the sentiment analysis. Many studies have analyzed twitter data for sentiment analysis and building on previous work, this article aims to identify the trends made by Twitter users related to the COVID-19 pandemic. Also, with the increase in the number of cases of COVID-19, analysis needs to be made about the fear factor in people. Due to the exponential rate of increment in the cases, many methods have been worked out to yield the precise result with less amount of time taken. In this article, we have used Sentiment Analysis in python using NLP tools like Bi-LSTM model and IBM Tone Analyzer.

## Literature Review

There have been many initiatives where NLP has been used to do sentiment analysis in various fields. Many models have been proposed to do the necessary. Earliest has been the use of naive Bayes Sentiment Classifier. The classifier takes a piece of text, from a document and transforms it into a vector of features with certain values ($f_1, f_2,..., f_n$). The classifier then computes the most likely sentiment polarity $S_j$, i.e. positive, negative, or neutral, given that we observed certain feature values in the text. In spite of being fast, it lacks accuracy with varied datasets. Deep Learning Models are then brought into the scene for the improvement in accuracy. Some models that have been tried out
.
for the Sentiment Analysis are Convolutional Neural Networks, Long Short Term Memory Networks and Recurrent Neural Networks. [3] uses Dirichlet allocation for topic modeling that is a generative statistical model allowing sets of observations to be explained. In [5], Machine Learning technique logistic regression classification method has been used along with the use of Naive Bayes method. But these have been constrained only to the country-specific tweets like Qatar [3] or USA [5]. We further extend the research on Indian Tweets with the modifications and collaboration of the used methods.

## Existing Approach

The idea of choosing the tone analyzer by IBM has been taken from [6] where the use of IBM Watson is done with the pairing of Google BERT Model. It uses the dataset from Twitter as well as two news channels, namely CNN and Fox News. In this approach, they have used TextBlob Polarity and Subjectivity Score. It is an NLP library where polarity score is calculated which is bipolar in nature, i.e., it generates scores in two forms, negative [-1] or positive [+1]. Subjectivity Score is also calculated in the two forms, Objective [0] and Subjective [+1].
This sums up manually labelling of the random sample data. Another way of labelling the tones in the random samples is the use of IBM Watson Tone Analyzer. It is a cloud service provided by IBM to do the sentiment analysis by dividing the tones in five broad categories. It does not confine the tone in two states, instead provides in-depth knowledge of the emotions. With the help of IBM Tone Analyzer, random samples of the dataset are labelled into tones, which then become the base for training data. So, the training data contains the output of manually labelling the data and IBM Watson Tone Analyzer. This is the training set for Google BERT Model, which is a machine learning algorithm for classification. BERT Model, which is a pretrained model, has been released in late 2018 which easily tunes the data to produce the desired output in the NLP area. It extracts high language features from the text and tunes them for classification. It uses less data and less time and yields good results. It already has its lower layers with initialized weights and addition of a single, untrained layer as the output layer and generally requires 2-4 epochs to run fully. Because of this advantage, it uses less dataset to produce the classification output with more accuracy. BERT consists of 12 Transformer layers. Each transformer takes in a list of token embeddings, and produces the same number of embeddings on the output.

For summarising the news articles' scraped data, LDA Modeling Technique has been used in the existing approach and has been classified into 8 topics which focuses on political and economical impact due to COVID-19. But the BERT Model, in spite of providing good results in less time, takes much of the complexity in understanding the model structure and applying the data on it for classification. We have tried to make our own classification model using Bi-LSTM, which can adapt to any dataset, with minimal complexity in making and understanding the model structure and its flow.

## Proposed Methods

### Data Collection

We collected coronavirus-related tweets between January 23, 2020, and May 20, 2020, using the web-scraping techniques. A total of 55 keywords were used in the web scraping technique to scrape out the tweets, which are the most currently used media terms relating to the new coronavirus outbreak. We extracted and stored the texts in .csv formats. Only English language tweets, that is, the tweets scripted in English, were collected in the study. Tweets on Twitter are retrieved with TwitterScraper API. For this, a Python Software Foundation is needed to be installed, namely the Tweepy library. The trending topics in this span of time are retrieved by scraping trendogate.com. These trends give us a broad spectrum of insight relating to COVID-19, which specifies what other activities were at peak despite the pandemic. The data retrieved can be put to use for many purposes. It can be used to study people's behavior, their interaction, opinions, and emotions during a specific time period. It can also show how people have coped up with the adversities. We have re-loaded the dataset(having recent dated tweets) at the end of the project due to the changing trends in the working time.

### Data Preprocessing

The research is initiated with the collection of Tweets. For the precise processing of the tweets for sentiment analysis, they are needed to be preprocessed. The preprocessing has multiple steps like data cleaning, data preparation, removal of special characters, and emoticons. Also, the URLs are eliminated from the raw data to make the text processable.

Initially, 1.9 lakh tweets have been scraped using TwitterScraper API and are sent for cleaning. In the cleaning of the data, English tweets are only selected, which are 1.2 lakh in number. In this lot, we also have those tweets that are scripted in English but are pronounced in Hindi language. They have a huge impact on the output as they are not correctly labelled. So, it poses a minor challenge to change them into the desired form. These sentences have been then converted to proper English sentences. These tweets are stored separately and are used further for sentiments' analysis.

For scraping trends in twitter, we have used a website called trendogate.com, which stores all the trendings since 2015. It stores worldwide twitter trends of each day, location-wise. We have scraped the website from the date 20th Jan 2020 to 20th May 2020.

*Data Analysis*

For analyzing the tweets, we have used IBM Watson Tone Analyzer. IBM Cloud provides the free service of the Tone Analyzer for the analysis of tones in the tweets. The free service is limited to a certain number of tweets. The Analyzer senses the emotional and language tones of each tweet and labels them into 7 categories namely **Fear, Anger, Tentative, Confident, Joy, Analytical, and Sadness**.

We submit JSON input that contains our written content, which is tweets in this context, to the service. The service accepts up to 128 KB of text, which is about 1000 sentences. The service returns JSON results that report the tone of our input. We store these results in the .csv file format and use them in improving the effectiveness of sentiment analysis. With the help of this, we are able to label the 85604 tweets, which become ready as the input to the Bi-LSTM Classification Model

*Model Summary*

As the block diagram above clearly shows our proposed model, Twitter data is extracted and sent for preprocessing. This data is fed to two places, to IBM Watson Tone Analyzer and to the Bi-LSTM Model for training and testing. The output of the Bi- LSTM Model then classifies the tweets under seven sentiments, which have been specified above. The Bi-LSTM Model has three hidden layers. First is the input layer, then the three hidden layers and then the output layer. For data to not overfit, a dropout layer has been added with the dropout rate as 0.5. First hidden layer has 128 cells, the second layer also has 128 cells and the third has 64 cells. The optimizer used throughout the Model is the ADAM Optimizer, which helps in updating the weight with less risk of having vanishing gradient problem. Loss function used here is the sparse categorical cross-entropy function. In the output layer, Softmax activation function has been used. All this has been implemented using Python 3, Keras.

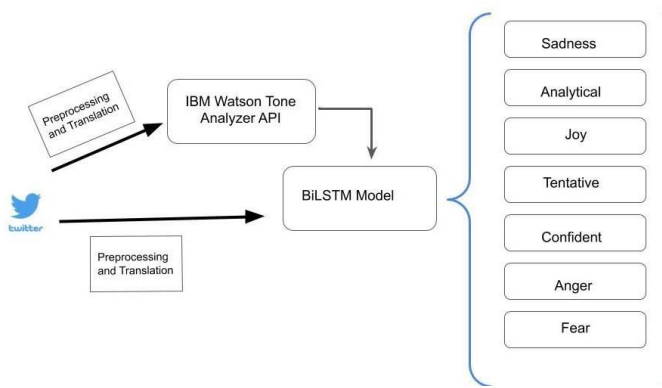

## Results

With the rigorous use of lakhs of tweets, Watson Tone Analyzer, we get to see the level of seven emotions in Figure 1. As clearly as we can see, people have been quite confident about surviving the epidemic and have shown less sign of fear and anger. We can also state that people were empathetic seeing the state of the world and the phase of quarantine brought a wave of sadness among them as they could not be with their loved ones. Though the sadness quota has been the prevalent emotion, confident sentiment has just been a few tweets behind it,

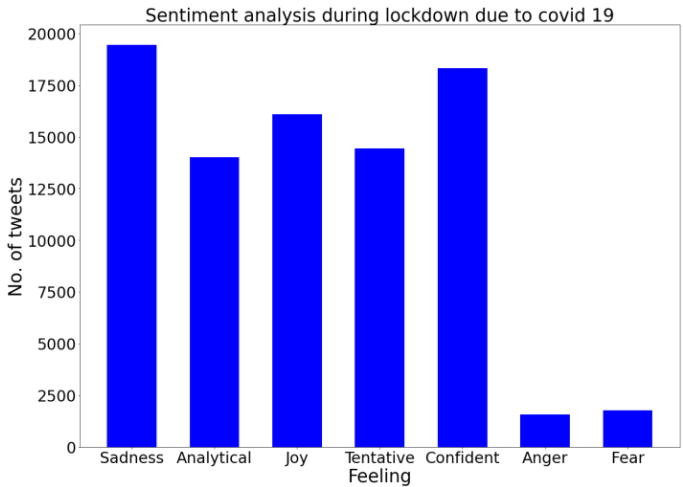

Figure 1. Sentiment Analysis

followed by Joy as the third most frequent emotion.

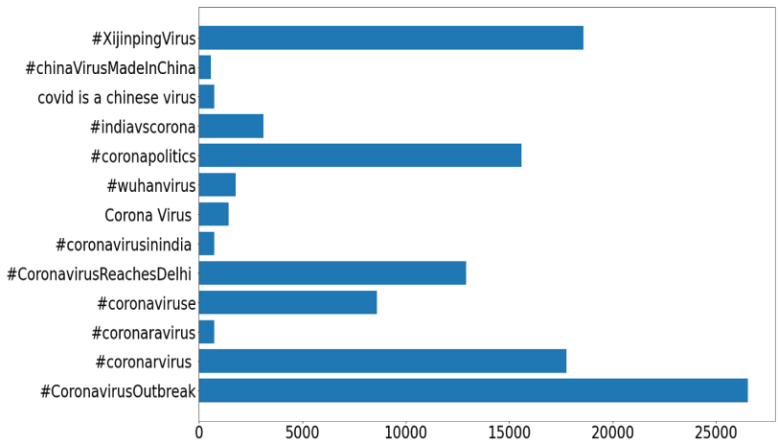

Figure 2. Trending Hashtags of COVID-19

In Figure 2, we see the trending hashtags in India while being in the lockdown situation due to the coronavirus. These are the top 13 trends versus the frequency of those hashtags. The most trending has been CoronaVirusOutbreak which is during the period of early March 2020.Also, while training the Bi-LSTM Mode on 6 epochs,loss comes out to be 0.45 and accuracy is at 86%.

The study shows the people of India have been very much positive throughout despite facing enormous challenges like population and diversity.

Table 1 shows the result of our overall model. Validation loss is seen to be 1.3899 whereas validation accuracy comes out to be 60% Now, while increasing the number of epochs to 10, loss reduces at 0.21 and accuracy increases to be 94%. Validation loss and accuracy are 1.65 and 59% respectively. When tested on this model, loss comes out to be 1.56 and testing accuracy to be 62%.

| Training | | |
|---|---|---|
| *Epochs* | *Loss* | *Accuracy* |
| 06 | 0.45 | 86 |
| 10 | 0.21 | 94 |
| **Validation** | | |
| 06 | 1.3899 | 60 |
| 10 | 1.65 | 59 |
| **Testing** | | |
| 10 | 1.56 | 62 |

*Table 1. Result*

Table 2 shows the precise report of each of the seven emotions that have been taken into consideration. Precision column indicates the fraction of relevant instances among the instances that have been totally retrieved. On the other hand, Recall column signifies the total amount of instances that were actually retrieved. $F_1$ score is a measure of statistical analysis that calculates a test's accuracy. The support is the number of samples of the true response that lie in that class. Anger and Fear have exclusively shown lower accuracy due to the lack of tweets that signified these two emotions. But the overall precision and accuracy surf to be more than 60%.

| Labels | Precision | Recall | f1-score | Support |
|---|---|---|---|---|
| 0 (Sadness) | 0.61 | 0.44 | 0.51 | 713 |
| 1 (Analytical) | 0.54 | 0.50 | 0.52 | 1109 |
| 2 (Joy) | 0.54 | 0.52 | 0.53 | 658 |
| 3 (Tentative) | 0.69 | 0.77 | 0.73 | 1538 |
| 4 (Confident) | 0.75 | 0.82 | 0.78 | 1567 |
| 5 (Anger) | 0.30 | 0.21 | 0.25 | 265 |
| 6 (Fear) | 0.30 | 0.41 | 0.34 | 214 |
| Accuracy | | | 0.63 | 6064 |
| Macro Average | 0.53 | 0.52 | 0.52 | 6064 |
| Weighted Average | 0.62 | 0.63 | 0.62 | 6060 |

*Table 2. Classification Report*

## Conclusion and Future Scope

The COVID-19 pandemic has been affecting many health care systems and nations, claiming the lives of many people. As a vibrant social media platform, Twitter projected this heavy toll through the interactions and posts people made related to COVID-19. Seeing the situation of the world, India knew the measures that were to be taken for fighting off the virus. Many came onto the conclusion that the virus certainly came from Wuhan, China and they made it vocal via all social media. Also, some took the advantage of the scene to sculpt it into the form of CoronaPolitics. As a conclusive result from the trends, one can say that people were highly active regarding the virus when it entered the peripherals of India. While some gave it the form of a political surrounding, many were enthusiastic about battling the virus against India and helping the nation survive. There have been many hurdles to overcome while working with the data. For instance, data cleaning took a lot of effort since India is known for its cultural and linguistic diversity and tweets were present in many languages which were hard to label under any particular sentiment. Moving forward, the percentage of sad and fear sentiments were less, due to which the training of Bi-LSTM Model was erratic in nature.

Nevertheless, we can extend it further for all the languages in the tweets to be processed . Also, more data can be extracted, forming a live feed, and giving the status of the sentiments each passing day.

## Conflicts of Interest

None declared.